\documentclass[10pt,twocolumn,letterpaper]{article}

\usepackage{iccv}
\usepackage{times}
\usepackage{epsfig}
\usepackage{graphicx}
\usepackage{amsmath}
\usepackage{amssymb}
\usepackage{booktabs}
\usepackage{xcolor}
\usepackage{epigraph}
\usepackage{multirow,float}
\usepackage{makecell}

\definecolor{lblue}{RGB}{187,222,251}
\definecolor{lpink}{RGB}{248,187,208}
\definecolor{yellow}{HTML}{EDD74C}
\definecolor{davered}{HTML}{f57c00}
\definecolor{lgreen}{HTML}{CCCCCC}
\definecolor{green}{RGB}{100,221,23}
\definecolor{darkpink}{HTML}{EC407A}

\usepackage[pagebackref=true,breaklinks=true,letterpaper=true,colorlinks,bookmarks=false]{hyperref}

\iccvfinalcopy 

\def\iccvPaperID{3128} 
\def\httilde{\mbox{\tt\raisebox{-.5ex}{\symbol{126}}}}

\newcommand{\argmin}[1]{\underset{#1}{\operatorname{arg}\,\operatorname{min}}\;}
\begin{document}

\title{Learning Temporal Dynamics from Cycles in Narrated Video}

\def\httilde{\mbox{\tt\raisebox{-.5ex}{\symbol{126}}}}
\newcommand{\de}[1]{{\color{red}[Dave: #1]}}
\newcommand{\model}[0]{MMCC}
\newcommand{\chen}[1]{{\color{blue}[Chen: #1]}}
\newcommand{\jw}[1]{{\color{cyan}[Jiajun: #1]}}
\newcommand{\cs}[1]{{\color{purple}[CS: #1]}}

\author{Dave Epstein\\
UC Berkeley$^*$
\and
Jiajun Wu\\
Stanford University
\and
Cordelia Schmid\\
Google
\and
Chen Sun\\
Google, Brown University
}

\maketitle
\ificcvfinal\thispagestyle{empty}\fi

\begin{abstract}
Learning to model how the world changes as time elapses has proven a challenging problem for the computer vision community. We introduce a self-supervised approach to this problem that solves a multi-modal temporal cycle consistency objective
jointly in vision and language. This objective requires a model to learn 
modality-agnostic functions to predict the future and past that undo each other when composed. We hypothesize that a model trained on this objective will discover long-term temporal dynamics in video. We verify this hypothesis by using the resultant visual representations and predictive models as-is to solve a variety of downstream tasks.
Our method outperforms state-of-the-art self-supervised video prediction methods on future action anticipation, temporal image ordering, and arrow-of-time classification tasks, without training on target datasets or their labels.
\end{abstract}

{\let\thefootnote\relax\footnote{{$^*$ Work done as an intern at Google Research.}}}

\section{Introduction}

\begin{figure}[t]
\centering
\includegraphics[width=\columnwidth]{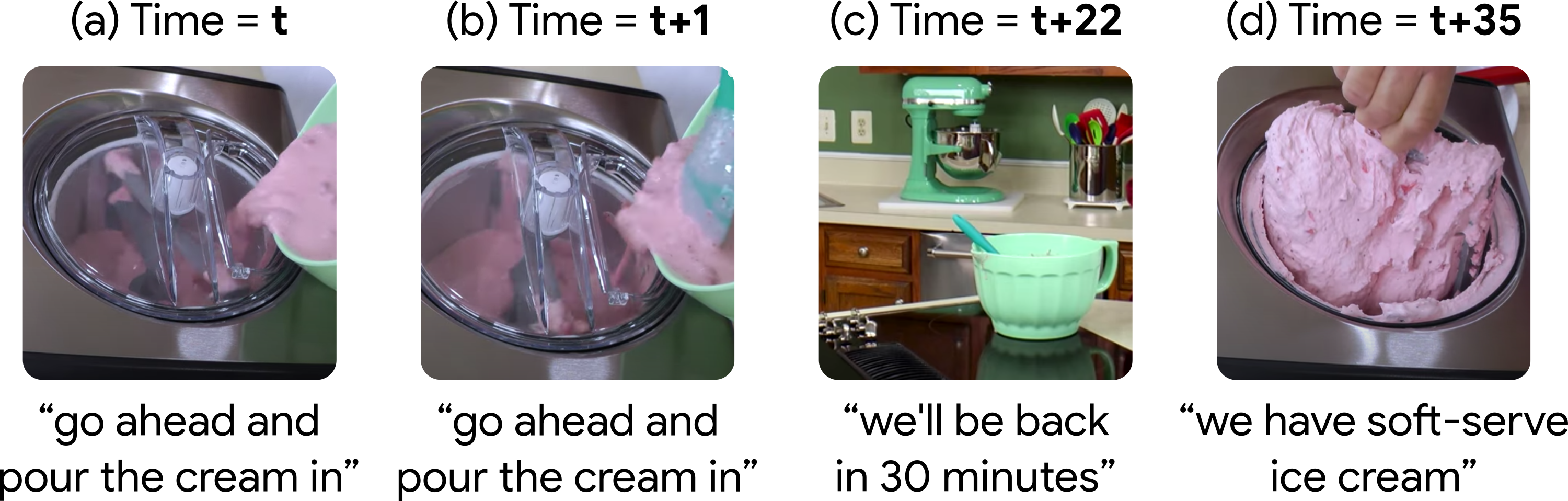}
\vspace{0.2em}
\caption{\textbf{Predicting the future is challenging.} Given a frame \textbf{(a)} at time $t$, previous work focuses on predicting frames at a fixed offset, such as \textbf{(b)}. However, these frames are often either redundant or stochastic, motivating the prediction of non-immediate futures. Predicting such a frame is highly non-trivial, as many are irrelevant, such as \textbf{(c)}. Aided by the textual information in narrated video, we can learn long-term temporal dynamics in video, and predict \textbf{(d)}. We learn these dynamics by solving a multi-modal temporal cycle consistency problem. }
\label{fig:teaser}
\end{figure}

Prediction is a central problem in computer vision which researchers have been grappling with since the early days of the field \cite{ebert2017self,finn2017deep,jayaraman2015learning,lee2018stochastic,mathieu2015deep,petrovic2006recursive,vondrick2016generating}. Previous deep learning methods have largely focused on predicting fixed, small offsets into the future. To understand why this formulation is flawed, consider Figure~\ref{fig:teaser}. This figure shows a frame \textbf{(a)} from a video at time $t$ and three frames at times $>t$. Which of the three should be the output of a model that predicts the future? Option \textbf{(d)} is closest to the future that humans are likely to imagine.
By predicting frames such as option \textbf{(b)}, which occur in the immediate future \cite{han2019video,han2020memory, sun2019learning,vondrick2016anticipating},
we limit the scope of temporal transitions that can be learned by models and hurt downstream performance.

Motivated by this example, we identify three central challenges in training a model to predict the future. First, manually annotating videos with temporal relationships between frames is prohibitively expensive, and ground truth may be difficult to define. Therefore, models should be able to learn from large unlabeled datasets of in-the-wild action and discover transitions autonomously, to enable practical applications. Second, modeling the complex long-term transitions in the real world requires learning high-level concepts, more naturally found in abstract latent representations than raw pixels. Finally, the duration elapsed by temporal transitions can vary significantly depending on context, and models must be able to make predictions at varied offsets into the future. To satisfy these desiderata, we introduce a new self-supervised training objective, \textbf{M}ulti-\textbf{M}odal Temporal \textbf{C}ycle \textbf{C}onsistency (\model{}), and a model that learns a representation to solve it. 



\begin{figure}[t]
\centering
\includegraphics[width=\columnwidth]{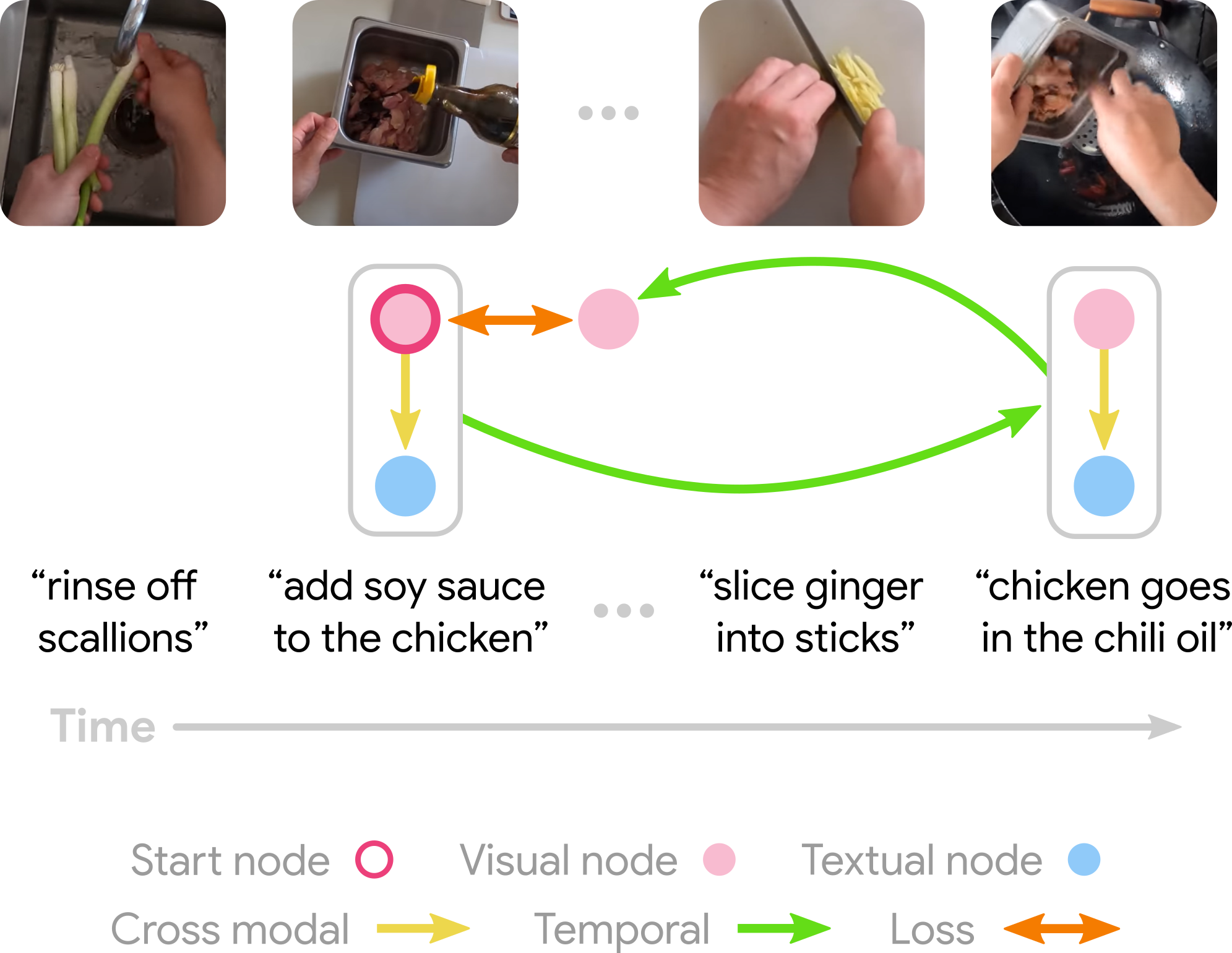}
\vspace{0.1em}
\caption{{%
\setlength{\fboxrule}{0.5pt}%
\setlength{\fboxsep}{1.5pt}%
\textbf{Learning temporal dynamics with cycles.} 
Given an \colorbox{lpink!50}{image} (here, second from left) as a start node,} our model finds corresponding \colorbox{lblue!50}{text} to build a \fcolorbox{lgreen}{white}{start state}. From this, our model predicts a future image and again builds a multi-modal state. Finally, our model predicts a past image from the future state. The discrepancy between this prediction of the past and the start image gives our \textcolor{davered}{\textbf{cycle-consistency loss}}. To solve this problem, we learn the \textcolor{green}{\textbf{temporal}} and \textcolor{yellow}{\textbf{cross-modal}} edges using soft attention. 
}
\label{fig:tempdyn} 
\vspace{-1.5em}
\end{figure}

 We show the \model{} objective in Figure \ref{fig:tempdyn}. Starting from a sampled frame in a narrated video, our model learns to attend among all narration text to retrieve a relevant utterance. Combining both modalities, the model learns a function to predict a latent future, attending over the entire video to retrieve a future frame. This frame's corresponding utterance is estimated, and a function to predict a past frame is learned in a similar way. The cycle constraint requires that the final model prediction be equal to the starting frame.  
 
 \model{} addresses all three challenges discussed above. In Figure~\ref{fig:teaser}, only \textbf{(d)} is a viable solution to our cycle formulation. Selecting \textbf{(c)} as a future would not allow the model to return to \textbf{(a)}, since the two frames have no clear relationship. On the other hand, because the model does not know which modality its input comes from---and therefore must operate equally on vision and language---it is discouraged from selecting lower-level future frames such as \textbf{(b)}, which likely do not accompany a predictable change in text.  
 

We show that our model, trained end-to-end from scratch to solve the \model{} objective on the HowTo100M dataset \cite{miech2019howto100m}, captures long-term dynamics in its predictive model of the future, and can be used without further training to anticipate future actions, order image collections, and identify salient temporal relationships in long videos. It also learns representations of video and text that contain information relevant to modeling temporal dynamics, which we demonstrate to be crucial to the quality of prediction.



Our main contributions are:

\begin{itemize}
    \item \model{}, a self-supervised multi-modal temporal cycle consistency objective that requires learning visual representations attuned to temporal dynamics, as well as long-term predictive models of the future and past.
    \item An attention-based model to solve this objective, which uses cross-modal and temporal cues to discover relationships through time in video.
    \item Since no previous self-supervised benchmarks exist in this area, a suite of qualitative and quantitative tasks to evaluate learned representations and predictive models. Our model outperforms the self-supervised SOTA in video prediction on all tasks.
\end{itemize}


\section{Related Work}
\noindent \textbf{Modeling the future.} Building predictive models of the future is a long-studied task in the computer vision community. Early work considers generating or warping pixels or optical flow to synthesize immediate futures~\cite{ babaeizadeh2017stochastic,finn2016unsupervised, lotter2016deep, pintea2014deja, ranzato2014video, srivastava2015unsupervised, vondrick2016generating,vondrick2017generating, walker2014patch, walker2015dense, yuen2010data}. More recent work attempts to model uncertainty in pixel-space, often by learning a distribution of futures that can be sampled~\cite{denton2018stochastic,hsieh2018learning,tap,kumar2019videoflow,lee2018stochastic,sun2019stochastic,villegas2017decomposing,villegas2017learning,xue2016visual}. These approaches tend to focus on synthetic or very short-term data, since synthesis is challenging in real video. Rather than predicting pixels, another line of work uses supervision to predict future action labels~\cite{huang2014action,kitani2012activity, lan2014hierarchical,miech2019leveraging, sadegh2017encouraging}. Sun~\etal~\cite{sun2019videobert} also uses narrated video, but quantizes input video using Kinetics supervision, then learns a transformer-based model of vision-and-language sequences. Instead of using supervision, Vondrick~\etal~\cite{vondrick2016anticipating} predicts {\em representations} which are trained to capture abstract concepts but are automatically obtained on large collections of data. Recent work extends this, using contrastive learning or other techniques to predict future representations~\cite{furnari2019would,han2019video,han2020memory, sun2019learning, wu2020learning}. With very few exceptions~\cite{tap}, this line of work is concerned with predicting time $t+1$ given time $t$. This formulation is highly constraining. Our model can predict arbitrarily far into the future and learns long-term dynamics from unlabeled, narrated video.

\noindent \textbf{Learning from unlabeled narrated video.} Self-supervised learning has a long history, even dating back to the early 1990s, where De Sa~\cite{desa} considered audiovisual data to ``derive label[s] from a co-occurring input to another modality''. We join an increasingly popular line of work and leverage automatic textual transcripts extracted from narrated videos uploaded online. Combining video and text has been widely explored in the deep learning era, with datasets largely focusing on manual textual annotation of video~\cite{didemo, epic, msrvtt, youcook} or on movies which have provided scripts~\cite{mpii, lsmdc}. Other work instead learns from automatic transcripts of narrations in instructional videos \cite{alayrac2016unsupervised,malmaud2015s,yu2014instructional}. A main benefit of learning from unlabeled video is that it unlocks unprecedented scales of data; Miech~\etal\cite{miech2019howto100m} introduces a dataset of over 100 million video clips and their narration transcripts, which is later used to learn strong models of cross-modal correspondence~\cite{milnce}. We are inspired by their success in training vision-and-language models on large collections of narrated video, and build on their data and approach to learn temporal dynamics.

\noindent \textbf{Learning with self-supervised cycles.} Cycle consistency was recently proposed \cite{zhou2016learning} as a natural cue for learning from unlabeled data or when ground truth is unavailable. In Zhu~\etal~\cite{zhu2017unpaired}, cycles are used for unpaired image-to-image translation; Recycle-GAN~\cite{Recycle-GAN} builds on this in follow-up work that incorporates simple temporal prediction (one timestep into the future) into these cycles. Kulkarni~\etal\cite{kulkarni2019canonical} uses cycles to learn mappings between canonical 3D surfaces and 2D images. Dwibedi~\etal\cite{dwibedi2019temporal} uses cycles to enforce that moments from two different videos should be mutual nearest neighbors, aligning action sequences and learning features useful for downstream tasks. Another line of work uses cycles to track objects through time~\cite{jabri2020space, wang2019learning}, tracking a pixel forward and then backward in time and requiring that the final pixel be the same as the start pixel. We are inspired by all these applications and introduce a new type of temporal cycle, one which not only incorporates multi-modal information into its learning, but also predicts dynamically into the future, instead of at a fixed offset. In particular, we draw inspiration from Jabri~\etal\cite{jabri2020space}, which casts temporal edges as contrastive comparisons (\ie, attention) among candidate nodes.

\section{Learning to Cycle through Narrated Video}

Our model learns long-term temporal dynamics by cycling through narrated video. We formulate the cycle consistency problem as follows: Given a moment $M_i$ in a start modality $M$ (either video $V$ or text $T$), retrieve a corresponding moment in the other modality $M^\prime$, then use both modalities to select a future moment in $M$. From this future moment, find a correspondence in $M^\prime$, then select a past moment in $M$. For the cycle to be complete, this final moment must be the same as the initial moment $M_i$. We illustrate the cycle in Figure~\ref{fig:tempdyn}. Solving this problem requires learning forward- and backward-in-time predictive functions that invert each other, as well as image and sentence embeddings that capture inter-modal correspondences and temporally relevant information.

\subsection{Cycles as repeated soft attention}

Let $V_{t_0:t_1}$ and $T_{t_0:t_1}$ be sequences of video and text, respectively, drawn from some temporal interval $[t_0, t_1]$. These sequences can be discretized into frames $\{V_i\}_{i=1}^{N_V}$ and utterances $\{T_i\}_{i=1}^{N_T}$, where $N_V,N_T$ are the number of instances the sequence is split into. We refer to each instance as a node, which allows viewing the training goal as learning a cyclic path through a graph, as depicted in Figure \ref{fig:tempdyn}. 

\begin{figure}[t]
\centering
\includegraphics[width=\columnwidth]{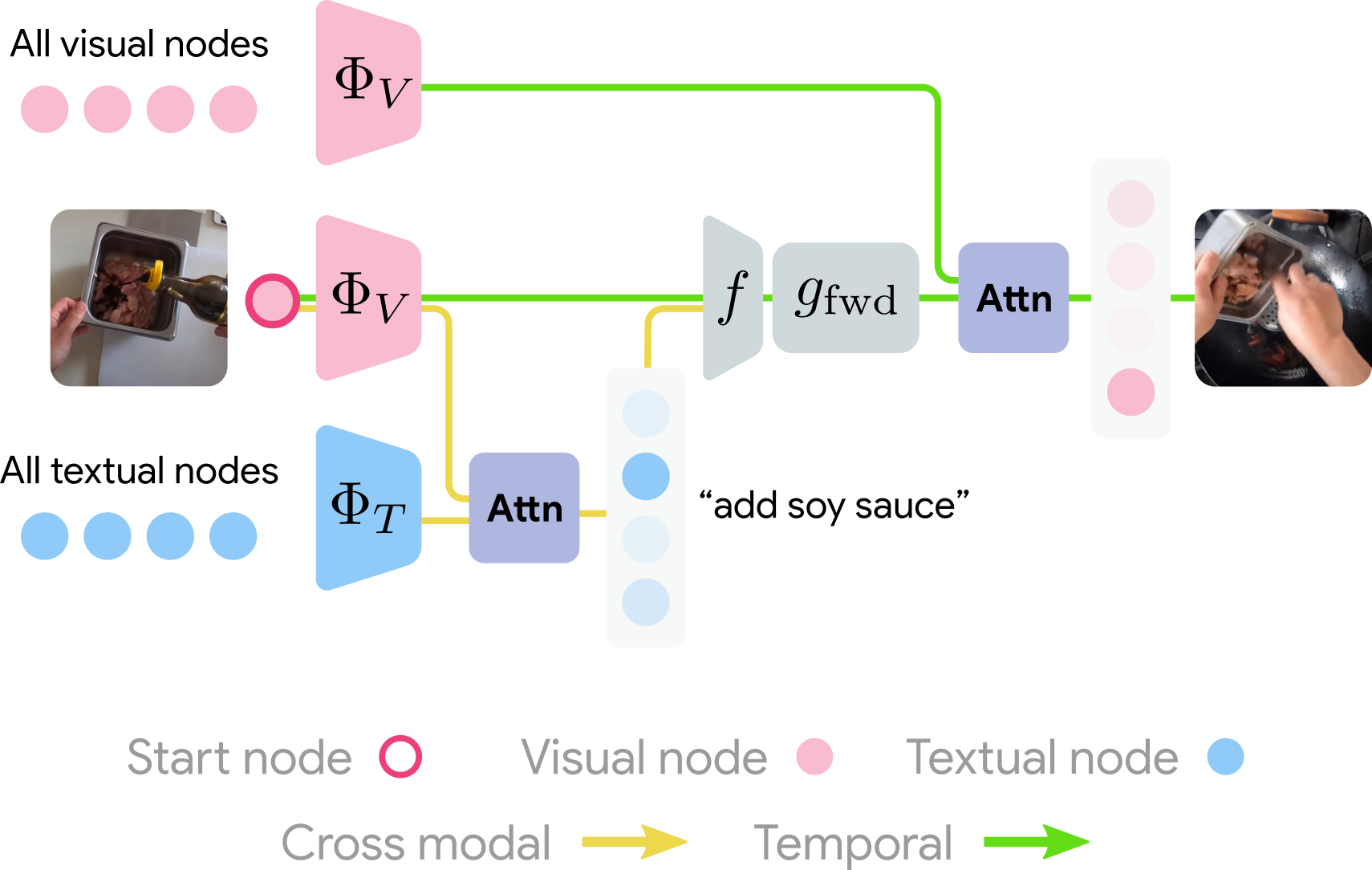}
\vspace{0.1em}
\caption{\textbf{Structure of a cycle edge:} 
\setlength{\fboxrule}{0.5pt}%
\setlength{\fboxsep}{2pt}%
We learn to embed all \colorbox{lpink!50}{visual} and \colorbox{lblue!50}{textual} nodes with $\Phi_V$ and $\Phi_T$. We then compute the \textcolor{yellow}{\textbf{cross-modal node}} corresponding to the start node with attention across the other modality. Both node representations are passed into an MLP $g_\text{fwd} \circ f$, which \textcolor{green}{\textbf{predicts the future}} using attention across the start modality. The process is then repeated to \textcolor{green}{\textbf{go backward in time}}, replacing $g_\text{fwd}$ with $g_\text{back}$. Our loss trains the model's final output to close the cycle by predicting the start node.}
\label{fig:breakdown} 
\end{figure}

In order to differentiate through the cycle generation process, let $uv$ be an edge in the graph shown in Figure \ref{fig:tempdyn}. We implement edges as soft retrievals of $v$ given $u$, as shown in Figure \ref{fig:breakdown}. This soft retrieval operation can be viewed as an application of the well-known attention mechanism \cite{attention}.

We start by running all visual and textual nodes through embedding networks $\Phi_T$ and $\Phi_V$ initialized with random weights. We use the architecture from \cite{milnce} for embedding text nodes and a ResNet-18 \cite{resnet} for visual nodes. This operation yields series of embeddings $\{z_{V_i}\}_{i=1}^{N_V}$ and $\{z_{T_i}\}_{i=1}^{N_T}$. 

We then compute the cycle edges, where each edge is an instance of soft attention as described above. The attention operation $\mathbf{Attn}(Q, K, V)$ accepts sets of query, key, and value vectors $Q \in \mathbb{R}^{N_Q \times d}, K, V \in \mathbb{R}^{N_K \times d}$ and returns a set of new values $Z  \in \mathbb{R}^{N_Q \times d}$, computed (with $\tau$-temperature softmax along the second dimension) as
\begin{equation}
    Z = \mathbf{Attn}\left(Q, K, V\right) = \text{Softmax}\left(\frac{Q K^T}{\tau}\right) V.
\end{equation}
To cycle through narrated video, we first select a modality $M \in \{T, V\}$ and a start node $M_\alpha \in M_{t_0:t_1}$ (we describe the process for selecting $M_\alpha$ in Section \ref{sec:startnodes}). We find the representation of the corresponding node in the other modality $M^\prime$ with a cross-modal attention edge:
\begin{subequations}
\begin{equation}
    z_{M^\prime_\alpha} = \mathbf{Attn}\left(\pi_{M_\alpha}, \{\pi_{M^\prime_i}\}_{i=1}^{N_{M^\prime}}, \{z_{M^\prime_i}\}_{i=1}^{N_{M^\prime}}\right).
\end{equation}
We learn to project representations $z$ into a shared semantic space, using a modality-specific projector $\Pi_M : \mathbb{R}^d \to \mathbb{R}^d$ that outputs vectors $\pi$. For notational convenience, we denote by $\mathbf{Attn}_{n_0:n_1}$ the same attention operation which considers keys (and values) at indices $\{n_0, \ldots, n_1\}$. In this notation we can rewrite the above as:
\begin{equation}
z_{M^\prime_\alpha} = \mathbf{Attn}_{1:N_{M^\prime}}\left(\pi_{M_\alpha}, \{\pi_{M^\prime_i}\}, \{z_{M^\prime_i}\}\right).
\label{eq:crossmodal}
\end{equation}
\end{subequations}

The representations from both modalities are concatenated and run through a multi-layer perceptron $f: \mathbb{R}^{2d} \to \mathbb{R}^d$. This operation yields $z_\alpha = f({z}_{M_\alpha}, z_{M^\prime_\alpha})$, embedding the joint information back into the shared semantic space. This choice also allows us to train our temporal edges without cross-modal information and accept input from only one modality with some probability $p_\text{unimodal}$, since $\Phi_V$ and $\Phi_T$ also map to $z$-space, in $\mathbb{R}^d$. 

Our model must now go from this multi-modal state representation $z_\alpha$ to a future state representation $z_\beta$. First, we predict an estimated representation of the future in projection ($\pi$) space, with an MLP $g_\text{fwd}$. We then retrieve the node in modality $M$ corresponding to this future state with a forward-in-time attention edge:
\begin{equation}
    z_{M_\beta} = \mathbf{Attn}_{1:N_M}\left(g_\text{fwd}(z_\alpha), \{\pi_{M_i}\}, \{z_{M_i}\} \right).
    \label{eq:fwd_attn}
\end{equation}
It is important 
to note that attention is order-invariant in $K$ and $V$, \ie shuffling rows of $K$ and $V$ yields the same output $Z$, since the individual matrix-row multiplications are agnostic to row index. This means that, importantly, {\em the model is not given temporal information about input nodes}, which could be used as a shortcut in learning\footnote{{\em E.g.}, the model could cycle by selecting the node it knows is at $t+1$ or $t-1$.}.
We then retrieve the corresponding node in $M^\prime$, $z_{M^\prime_\beta}$, with another cross-modal edge, projecting queries and keys into $\pi$-space:
\begin{equation}
    z_{M^\prime_\beta} = \mathbf{Attn}_{1:N_{M^\prime}}\left(\pi_{M_\beta}, \{\pi_{M^\prime_i}\}, \{z_{M^\prime_i}\}\right).
    \label{eq:crossmodalbck}
\end{equation}
As before, these vectors are combined to yield a future state representation $z_\beta = f(z_{M_\beta}, z_{M^\prime_\beta})$.

This process is repeated to predict backward in time. We compute $g_\text{back}(z_\beta)$, where $g_\text{back}$ shares its first few layers
with $g_\text{fwd}$ to allow learning features useful for dynamics in either direction (see Section \ref{sec:impl} for more details). To close the cycle, we compute the normalized similarity scores between $g_\text{back}(z_\beta)$ and the $\pi$-space nodes in $M$:
\begin{equation}
    \mathbf{p} = \mathbf{Attn}_{1:N_M}\left(g_\text{back}(z_\beta), \{\pi_{M_i}\}, \{1\}  \right).
    \label{eq:back_attn}
\end{equation}
We train our system with the negative log likelihood loss on the score vector cycling back to the location of $M_\alpha$, which we denote $i_\alpha$:
\begin{equation}
\label{eq:loss}
    \mathcal{L}_\text{cycle} = - \log \mathbf{p}^{(i_\alpha)}.
\end{equation}

\subsection{Cross-modal correspondence}
\label{sec:methodxm}

A key component of our cycle model is the ability to find correspondences between vision and language. Eqs.~\ref{eq:crossmodal} and \ref{eq:crossmodalbck} crucially rely on this ability in order to incorporate multi-modal information into temporal edges. Recent work has demonstrated remarkable progress in training models on massive datasets for this cross-modal retrieval task: given a moment at time $t$ in one modality of a video - $M_t$ - find the matching moment in the other modality $M^\prime$. 

We build on the approach presented in \cite{milnce} which uses a contrastive loss to train representations of vision and language, where temporally co-occurring information is considered ground truth ($M_t$ should retrieve $M^\prime_t$), and other vision-language pairs are used as negatives. To handle the common misalignment intrinsic to real-world video, \cite{milnce} allows for representations within $k$ nodes of the ground truth node $M^\prime_t$ to be considered as positives. We adopt this approach to learn cross-modal correspondence, training it for finer-grained discrimination among a set of candidate moments drawn from the same video as opposed to randomly across the entire dataset. We denote the loss used to train cross-modal correspondence $\mathcal{L}_\text{cross-modal}$. For the full cross-modal formulation, please see Supplementary Material. 

\subsection{Starting the cycle}
\label{sec:startnodes}
Our model will be unable to learn semantic transitions between states if the initial input node depicts noisy or unclear data. This is especially probable when training on unconstrained, real-world video datasets. Therefore, instead of randomly sampling start nodes, we sample from a distribution defined by a ``concreteness'' score $s_i$. We calculate this score for each node $M_i \in M$ as the highest cross-modal similarity between $M_i$ and some node $M^\prime_j$ in the other modality. Intuitively, this score captures concreteness since frames and utterances that align strongly tend to contain objects or actions which are salient in both modalities:
\begin{equation}
    s_{i} = \max_j \left( \pi_{M_i} \cdot \pi_{M^\prime_j} \right)
    \label{eq:concreteness_cos_prior}
\end{equation}


We run the above scores through a softmax with $\tau = 0.1$, yielding a distribution from which we sample $M_\alpha$. 

\subsection{Avoiding collapse}
Training on the above formulation of $\mathcal{L}_\text{cycle}$ in practice may lead to fast collapse to a simple ``looping in place'' solution, where temporal edges always point to the current node. We propose two strategies to prevent this collapse:

\noindent \textbf{Constraining candidate nodes.} We can limit the range of temporal edges during training by removing nodes from $K$ and $V$ in Eqs.~\ref{eq:fwd_attn} and~\ref{eq:back_attn}. We rewrite Eq.~\ref{eq:fwd_attn} with $\mathbf{Attn}_{i_\alpha + 1:N_M}$, \ie, since we know the index the cycle starts from, we can consider only those nodes after the start point in the forward edge. We similarly rewrite Eq.~\ref{eq:back_attn} with $\mathbf{Attn}_{1:\texttt{index}(z_{M_\beta}) - 1}$, where $\texttt{index}(z_{M_\beta}) = \arg \max_i \left( g_\text{fwd}(z_\alpha) \cdot \pi_{M_i} \right)$, \ie, the index of the node with highest similarity to the latent predicted future. This constrains the backward edge to only consider nodes that precede the estimated current index. This can also be seen as resolving the sign ambiguity inherent to the unconstrained formulation which allows the model to go back-then-forward or vice versa. Importantly, we run the model {\em without} this constraint at test time. 


\noindent \textbf{Penalizing visual similarity.} Alternatively, we can encourage our model to select visually diverse nodes in its temporal edges:
\begin{align}
    \mathcal{L}_\text{sim} &= \max \left( \cos(z_{V_\alpha}, z_{V_\beta}) - m ,0 \right) 
    \\ &+ \max \left( \cos(z_{V_\beta}, z_{V_{\alpha,\text{back}}}) - m ,0 \right)\nonumber,
\end{align}
where $z_{V_{\alpha,\text{back}}}$ is the visual representation given by Eq.~\ref{eq:back_attn}, replacing the values with $\{z_{M_i}\}$, and $m=0.5$ is a margin.

In practice, we combine both the above strategies for the strongest results.

\subsection{Implementation}
\label{sec:impl}
We combine $\mathcal{L}_\text{cycle}$, $\mathcal{L}_\text{cross-modal}$, and $\mathcal{L}_\text{sim}$ in our final loss:
\begin{equation}
    \mathcal{L} = \lambda_1 \mathcal{L}_\text{cycle} + \lambda_2 \mathcal{L}_\text{cross-modal} + \lambda_3 \mathcal{L}_\text{sim}.
\end{equation}

We embed images into $\mathbb{R}^d$ ($d=512$) using a ResNet-18 \cite{resnet}, and embed text using a word embedding matrix followed an MLP and global pooling, as in \cite{milnce}. 

We implement all modules ($\Pi_M$,  $f$, $g_\text{fwd}$, $g_\text{back}$) as MLPs, where each layer is followed by ReLU and a LayerNorm except for the final layer, which is followed by $l_2$ normalization if its output is in $\pi$-space. $\Pi_M$ and $f$ are one-layer MLPs, and $g_\text{fwd}, g_\text{back}$ are four-layer MLPs, with weights of the first two layers shared.
We randomly sample batches of video segments of maximum duration $t_1 - t_0 = 64 \text{sec}$. The sparsity at which data is sampled affects the time elapsed by input videos in a batch as well as the granularity of visual information provided to the model. Denser data is less likely to miss key moments, but more likely to contain redundant information. We therefore train models on various image sampling frame rates $r \in \{0.25 \text{fps}, 0.5 \text{fps}, 1 \text{fps}\}$.

Because good cross-modal correspondence is necessary to learn strong, semantic cycles, we initialize $\lambda_2 = 1$ and exponentially increase $\lambda_1$ from some small value $\epsilon$ up to $1$, across 30 epochs. We peg $\lambda_3 = 3 \lambda_1$ when using the similarity loss. For further details on training and architecture, please see Supplementary Material.

\section{Experiments}
This sections examines the design choices and learned temporal dynamics of our model. Since most previous benchmarks focus on supervised action anticipation with fixed categories and time offsets~\cite{epic,breakfast}, we design a suite of qualitative and quantitative experiments to evaluate different approaches.
\label{sec:experiments}
\subsection{Data}

We train our model on unconstrained real-world video data. Specifically, we use a subset of the HowTo100M dataset~\cite{miech2019howto100m}, which contains around 1.23 million videos and their automatically extracted audio transcripts. Videos in this dataset are roughly categorized by subject area, and we use only the videos categorized ``Recipe'', around a quarter of the dataset. We build a train-validation-test split such that of 338,033 total recipe videos, 80\% are in train, 15\% in validation, and 5\% in test. Recipe videos are rich in complex objects, actions, and state transitions, and the subset allows us to train models faster. 

For more controlled testing, we use the CrossTask dataset~\cite{crosstask}, which contains similar videos along with task-specific annotations. Videos are associated with tasks (\eg, ``making pancakes''), where each task has a predefined sequence of high-level subtasks with rich long-term temporal inter-dependencies (\eg, [``pour flour into bowl'', ``crack egg into bowl'', ..., ``drizzle maple syrup'']). Video segments that depict one of these subtasks are annotated as such. 

\subsection{Previous work and baselines}
\noindent
\textbf{Baselines:} We evaluate purely cross-modal features (Section \ref{sec:methodxm}), given by frozen embedding nets $\Psi_V$, $\Psi_T$, and also use these features as prediction targets for RA and TAP below. We also study ImageNet supervised features~\cite{imagenet}.

\noindent
\textbf{Representation Anticipation (RA):} 
As a representative of the self-supervised line of work in predicting a fixed offset into the future, we implement RA~\cite{vondrick2016anticipating} on our data and architecture, training a model to predict frozen representations of a network trained for cross-modal correspondence. In vision, we train the network to anticipate one second into the future, while in text, we anticipate the subsequent utterance (on average,  $\sim$2 seconds into the future). We train:
\begin{equation}
    \argmin{\Phi_V, \Phi_T, f_{t+1}} -\cos\Big(f_{t+1}(\Phi_M(M_i)), \Psi_M(M_{i+1}) \Big),
    \end{equation}

\noindent
\textbf{Time-Agnostic Prediction (TAP):} Noting the restrictive nature of the fixed offset formulation, TAP~\cite{tap} introduces the minimum-across-time formulation to allow the prediction of ``bottleneck'' predictable moments. We implement their loss, taking the minimum across all future moments in the sampled video segment:
\begin{equation}
    \argmin{\Phi_V, \Phi_T, f_{t+1}} \min_{i < i^\prime \leq N_M} -\cos\left(f_{t+\Delta}(\Phi_M(M_i)), \Psi_M(M_{i^\prime}) \right).
    \end{equation}
While the above two models do not consider the exact same setting as us, we re-implement their approaches as faithfully as possible, training them to predict SOTA features trained for cross-modal correspondence.

\noindent
\textbf{MemDPC:} In order to efficiently model multiple future hypotheses, MemDPC~\cite{han2020memory} casts future prediction as estimation of convex combinations of memories stored in a codebook, and achieves SOTA performance on tasks of interest. We evaluate their trained visual future prediction model, which does not take textual information as input.



\subsection{Evaluating cycle consistency}
\label{sec:variants}
Central to our formulation is the model's ability to learn dynamic predictions of the future and past that undo each other, as well as finding strong cross-modal correspondences. Thus, we begin by evaluating how well different model variants are able to solve our self-supervised objective on the Recipes test set. We ablate various design choices, including multi-modal information usage, cycle edge order, and temporal constraints on edges. 



\begin{table}[tb]
\centering
\small
\resizebox{\columnwidth}{!}{  
\begin{tabular}{l l | c c}
     \toprule
         \multirow{2}{*}{\textbf{Choice}}  & \multirow{2}{*}{\textbf{Variant}} & \multicolumn{2}{c}{\textbf{Percentile rank}}\\
     & & \textbf{Cycle} & \textbf{Cross-modal} \\ \midrule
     Temporal constraint & None* & - & - \\ 
     &Similarity loss & 93.1 & 74.4 \\
     &Max-index & 92.6 & 74.3 \\
     &\textbf{Max-index + sim. loss} & \textbf{93.6} & \textbf{75.7} \\
     \midrule
     Multi-modal info. & $p_\text{unimodal} = 1$ & 89.8 & 74.3 \\
    &$\mathbf{p_\textbf{unimodal} = 0.5}$ & \textbf{93.6} & \textbf{75.7} \\
    & $p_\text{unimodal} = 0$ & 96.5 & 75.9 \\ \midrule
    Start point selection & \textbf{Cross-modal similarity} & \textbf{93.6} & \textbf{75.7} \\
     &Random & 88.7 & 74.5 \\
    \midrule
     Input embedding & \textbf{Fine-tuned} & \textbf{93.6} & \textbf{75.7} \\
     & Frozen cross-modal \cite{milnce} & 67.5 & 76.8 \\
    \midrule
    Cycle path & \textbf{Within modalities} & \textbf{93.6} & \textbf{75.7} \\
    & Across modalities & 85.0 & 73.2 \\ \midrule
    & Chance & 50.0 & 50.0 \\
     \bottomrule
\end{tabular}
}
\vspace{0.2cm}
\caption{\textbf{Cycle-back and cross-model accuracy:} We evaluate models on the percentile rank assigned to ground truth in the cycle and cross-modal tasks on the Recipes test set (100 = ground truth ranked first, 0 = ranked last). Used options are shown \textbf{in bold}. *Without any temporal constraint, training collapses.}
\label{tab:variants}
\vspace{-1.5em}
\end{table}

\noindent
\textbf{Multi-modal information: } As an alternative to defining the state as a learned combination of visual and textual representations $z_\alpha = f({z}_{M,\alpha}, z_{M^\prime,\alpha})$, we can use only one modality at a time, giving $z_\alpha = z_{M,\alpha}$. The frequency at which only unimodal information is used can be controlled by a hyperparameter $p_\text{unimodal}$. 

\noindent
\textbf{Cycle path: } The above formulation navigates between moments in the start modality $M$, optionally using information from $M^\prime$ to augment representations. We denote this variant \textbf{Within modalities}. The order of these edges can also be permuted, such that cycles start in $M$, retrieve a moment in $M^\prime$, find a future moment in $M^\prime$, then cycle back through $M$. This variant is denoted \textbf{Across modalities}.

\noindent

\noindent
\textbf{Evaluating variants:} To compare between different variants, we measure the average percentile rank (\eg 100 = ground truth is ranked first among all candidates, 50 = ranked in the middle, 0 = ranked last) assigned by our model to ground truth cross-modal and cycle nodes. 
We show this ablation study in Table~\ref{tab:variants}, observing significant gains using our cycle configuration. We hypothesize that across-modality cycles perform worse since switching modalities acts as a bottleneck, forcing the model to discard information that would be useful for subsequent edges. 

\begin{figure}[t]
\centering
\includegraphics[width=\columnwidth]{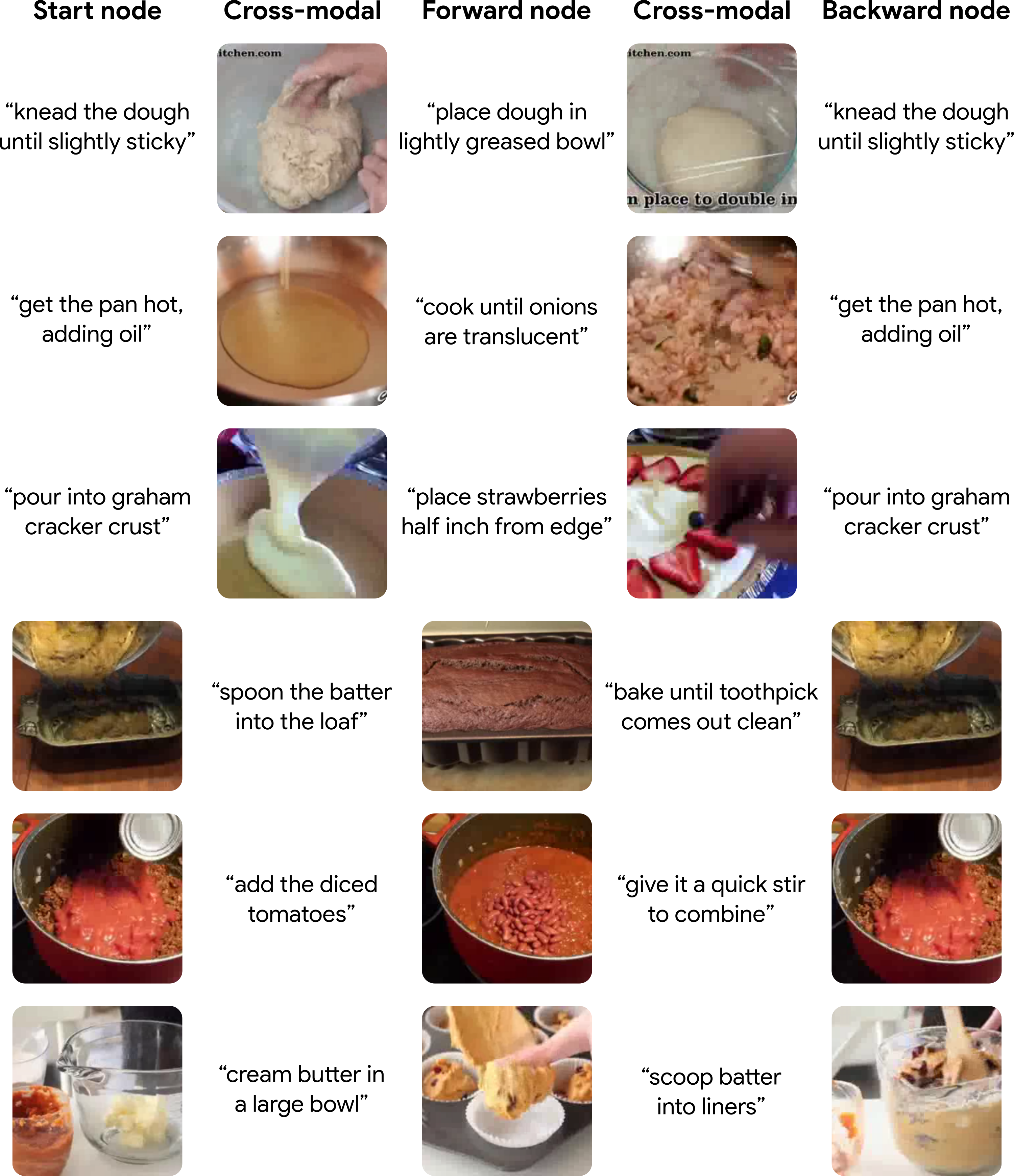}
\caption{\textbf{Emergent long-term temporal dynamics:} We show examples of learned model cycles in the Recipes test set. Given a start node (top: text, bottom: image) sampled as described in Section \ref{sec:startnodes}, we show the retrieved cross-modal node, the predicted future node and its cross-modal retrieval, and the model's final backward prediction. In the bottom row, we show a failure case, where the forward edge skips too far ahead and breaks the cycle.}
\label{fig:cycle_examples} 
\end{figure}

\noindent
\textbf{Visualizing cycles:} We show examples of cycles discovered by the trained model in Figure \ref{fig:cycle_examples}. Our model correctly cycles back around 66\% of the time (chance is 4\%). The model appears to traverse video according to long-term dynamics, as hypothesized. Note that these transitions occur up to one minute apart, highlighting the importance of allowing dynamic prediction offsets. 

\subsection{Zero-shot prediction}


\begin{figure}[t]
\centering
\includegraphics[width=0.85\columnwidth]{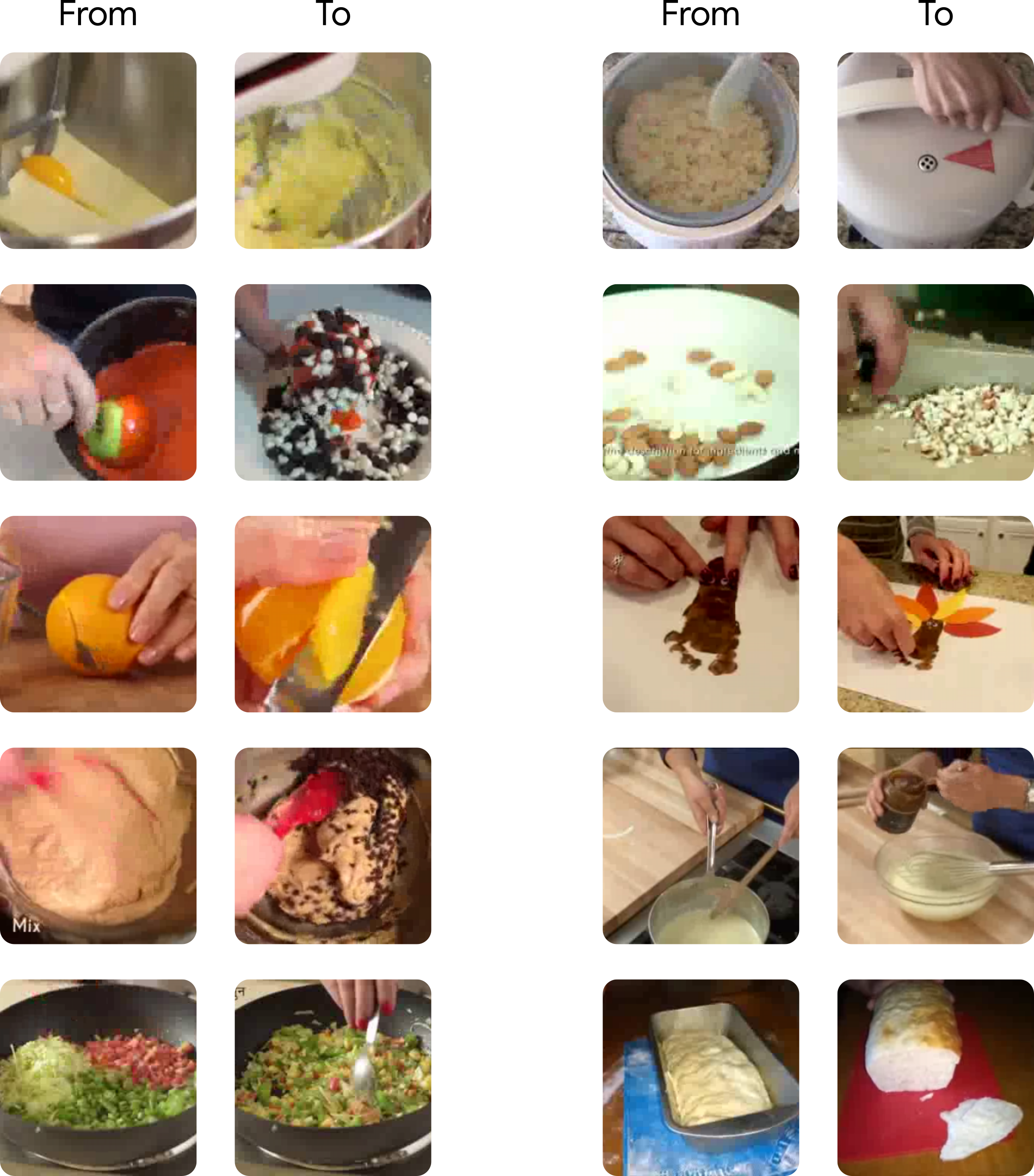}
\vspace{0.3em}
\caption{\textbf{Discovering transitions in video:} Once trained, the learned model of temporal dynamics can be applied to long (5-10 min.) video sequences to discover the most salient pairwise transitions $u \to v$. We compute the probability of a transition as defined in Eq.~\ref{eq:putov} and show the highest-score transitions in 10 different Recipes test set videos. }
\label{fig:transition_examples} 
\end{figure}

\noindent
\textbf{Ranking transitions by likelihood}: To directly evaluate the learned representations $\pi$ and functions $g_\text{fwd}$ and $g_\text{back}$, we can visualize the pairs of frames $(u, v)$ for which the probability of $v$ being the future of $u$ is highest. We model this probability as the product of the likelihood of states $u$ and $v$ and the forward and backward likelihood of $u \to v$:
\begin{align}
\label{eq:putov}
    P_\text{fwd}(v | u) &= \frac{e^{\pi_\text{fwd} \cdot \pi_v}}{\sum_{m \in M} e^{\pi_\text{fwd} \cdot \pi_m}},  \\ \nonumber
    P_\text{back}(u | v) &= \frac{e^{\pi_\text{back} \cdot \pi_u}}{\sum_{m \in M} e^{\pi_\text{back} \cdot \pi_m}}, \\ \nonumber
    P(u \to v) &= P_\text{fwd}(v | u) \cdot P_\text{back}(u | v) \cdot P(u) \cdot P(v),
    \label{eq:utov}
\end{align}
where $\pi_\text{fwd},\pi_\text{back}$ are the result of running $u$ and $v$ (optionally with cross-modal information) through $g_\text{fwd}, g_\text{back}$ and $P(x)$ is the concreteness score defined in Equation \ref{eq:concreteness_cos_prior}.

We compute this probability efficiently for all $n^2$ pairs in long clips of continuously sampled video ($n \approx 600$). We then look at the top temporal transitions discovered by the model in each video. We show results on the Recipes test set in Figure \ref{fig:transition_examples}. The top transitions show clear state transitions such as adding chocolate chips to dough, segmenting an orange, and baking a loaf. These predictions could not be made by a model trained to predict a fixed future, since they occur at varied temporal offsets.

\noindent
\textbf{Predicting future actions:} Existing benchmarks {\em e.g.} \cite{epic,breakfast} focus on predicting action from a few frames in the immediate past or present. Instead, given a few frames, we wish to predict long-term temporal dynamics, which may unfold arbitrarily far into the future. While the former task is more well-defined, the latter is more interesting and relevant. However, ground truth for this task -- \ie, per-frame annotation of related future action -- is not widely available. We propose using CrossTask task steps as a proxy, since they capture long-term temporal relationships in video. 


\begin{table}[tb]
  \centering
  \resizebox{\columnwidth}{!}{  
\small
\begin{tabular}{l | c cc| c  c c}
     \toprule
     \multirow{2}{*}{\textbf{Model}}& \multicolumn{3}{c|}{\textbf{Recall}}& \multicolumn{3}{c}{\textbf{Percentile rank}}\\
     &\textbf{@ 1} & \textbf{@ 5} & \textbf{@ 10}& \textbf{Worst} & \textbf{Mean} & \textbf{Best} \\ \midrule
     MemDPC{*} \cite{han2020memory} & 2.9 & 15.8 & 27.4 & 25.6 & 48.4 & 71.4 \\
     Cross-modal \cite{milnce} & 2.9 & 14.2 & 24.3 & 28.2 & 47.9 & 68.2 \\
     Repr. Ant. \cite{vondrick2016anticipating} & 3.0 & 13.3 & 26.0 & 25.7 & 47.7 & 71.4 \\
     TAP \cite{tap}  & 4.5 & 17.1 & 27.9 & 28.3 & 50.1 & 71.6\\
     \textbf{\model{} (ours)} & \textbf{5.4} & \textbf{19.9} & \textbf{33.8} & \textbf{33.0} & \textbf{55.0} & \textbf{76.9} \\ 
     \bottomrule
\end{tabular}
}
\vspace{0.2cm}
\caption{\textbf{Predicting future actions:} We evaluate models' ability to anticipate action at a high level, potentially minutes into the future, without any fine-tuning. On the CrossTask dataset \cite{crosstask}, our model outperforms the previous self-supervised state of the art in inferring possible future actions. *We evaluate MemDPC by clip retrieval since it does not have a textual representation.} 
\label{tbl:xtask_pred}
\vspace{-1em}
\end{table}

For a video $V$ belonging to task $\tau$ (with $N_\tau$ predefined subtasks), let $v \in V$ be a clip with subtask label $T_{\tau, i}$ ($i^\text{th}$ in the predefined sequence). We would like to predict future actions $T_\text{future} = \{T_{\tau, j}\}_{j=i+1}^{N_\tau}$ from $v$. For example, given a short clip of eggs being added to a bowl with flour, the model should assign high likelihoods to subtasks such as ``mix batter'' and low likelihoods to ``crack eggs'' or ``season steak''. Formally, we define a future likelihood score given a video segment $v$ and candidate future subtask $T_j$. We first sample frames from the video segment and compute their average embedding $\bar{z}_{v}$. The likelihood score uses our learned representations and predictive model to define a score $s_{v \to j} = g_\text{fwd}(\bar{z}_{v}) \cdot \pi_{T_j}$. We compute likelihood scores for all subtask descriptions in the CrossTask validation set, and consider the model's prediction correct if any of the future actions in $T_\text{future}$ are predicted, since not all future subtasks are necessarily related to the given visual state.

Table \ref{tbl:xtask_pred} shows recall and percentile rank statistics for this task. We compare our model to \cite{han2020memory,tap, vondrick2016anticipating}, replacing $g_\text{fwd}$ with each method's predictive model. Since \cite{han2020memory} is vision-only, we set $\pi_{T_j}$ to the average visual representation of all video segments with a given subtask label $T_j$. We also define a cross-modal similarity score $s_{v \to j} = \bar{\pi}_{v} \cdot \pi_{T_j}$ as a strong baseline, taking advantage of contextual similarities in video and text. 
Our model outperforms all baselines and self-supervised state of the art on detecting the temporal relationships between visual states and future actions. 

\subsection{Further analysis}
\begin{figure}[t]
\centering
\includegraphics[width=0.85\columnwidth]{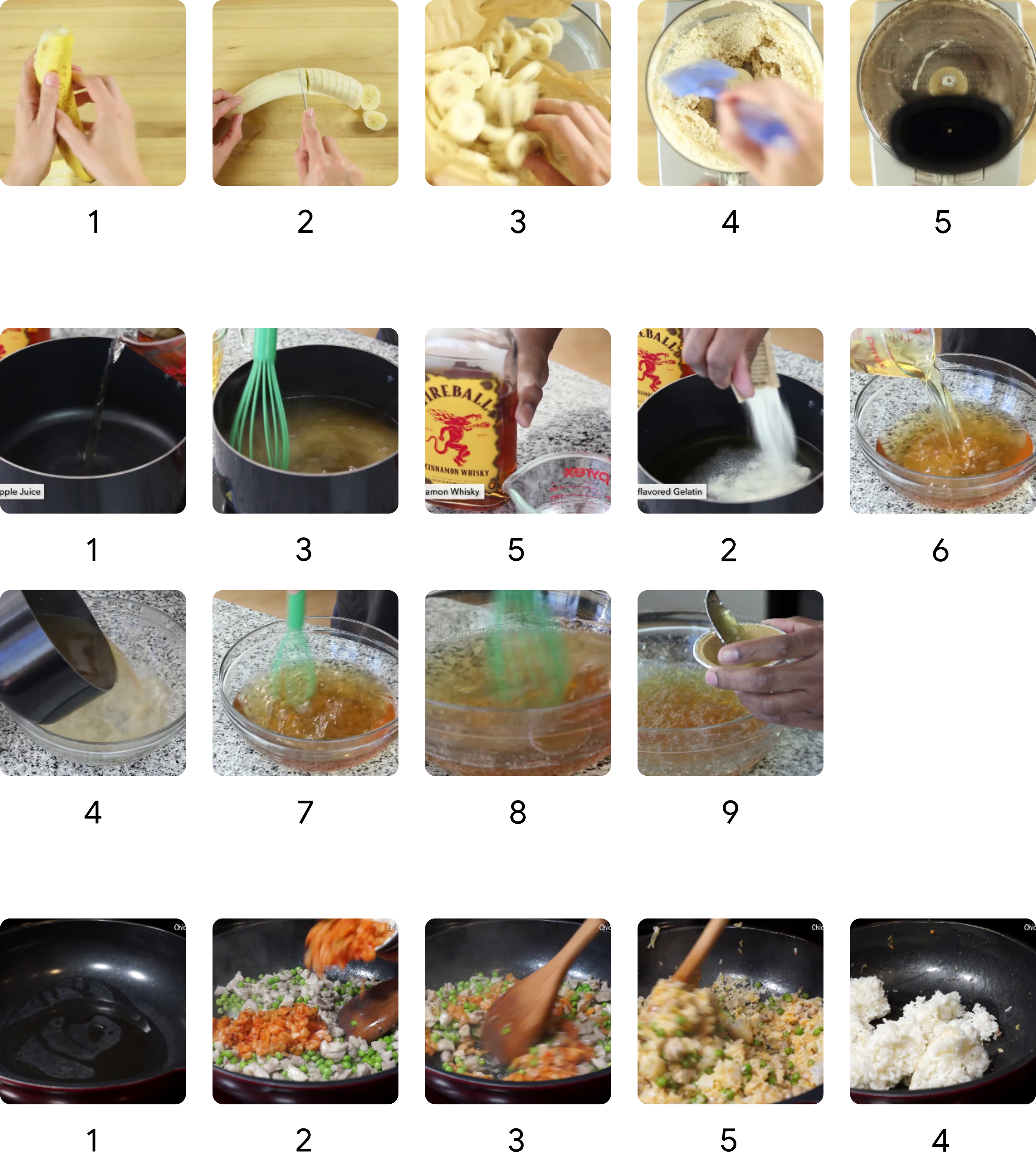}
\vspace{0.2cm}
\caption{\textbf{Unshuffling image collections:} We show example video sequences in the ordering given by treating the induced graph of log-likelihoods as an instance of the traveling salesperson problem. We list the ground truth index in the sequence under each clip. Even accepting only sparse video frames as input, our model makes reasonable predictions on this challenging task.}
\label{fig:unshuf_examples} 
\end{figure}

\noindent
\textbf{Unshuffling bags of frames:} The ability to order a shuffled set of states is used to evaluate human language and common-sense reasoning skills, and has been explored as a learning signal in NLP \cite{gong2016end, lee2020slm, logeswaran2016sentence}. This same ability can also be used to discover temporal structure and summaries of events from large image datasets, as in \cite{kim2014joint}. We solve this problem by finding the optimal explanation of shuffled video given by iterative application of our temporal dynamics model. Out of all $n!$ possible orderings $\{i_1, i_2, \ldots, i_n\}$, we select the one for which $\prod_j P(x_{i_j} \to x_{i_j+1})$ is highest.

Given $P(u \to v)$ scores computed by Eq.~\ref{eq:putov}, we induce a fully-connected directed graph with sampled frames as nodes and edge weights given by $\texttt{weight}(\vec{uv}) \equiv -\log P(u \to v)$. Adding a special null node connected to all other nodes with edge weight 0 allows running this graph through an off-the-shelf traveling salesperson problem (TSP) solver\footnote{https://pypi.org/project/elkai/}. The optimal TSP solution then represents the lowest-cost (ordered) path through all video clips, effectively unshuffling the input.

We run this experiment on CrossTask, where videos are annotated with ordered steps and their associated temporal segments. We treat each segment as a node by computing its average visual representation, as before. We then use these representations to find $P(u \to v)$ scores between labeled segments and solve an optimal path. We run this experiment both in vision only (by passing projected visual representations directly into $g$ in Eq.~\ref{eq:putov}) as well as with ground truth vision-text pairings, and show results in Table~\ref{tbl:deshuf}. We show example predicted orderings in Figure \ref{fig:unshuf_examples}. Again, we can replace $g_\text{fwd}$ with the future prediction model in other methods and run the same algorithm. Our model outperforms previous work on all evaluation metrics.

\begin{table}[tb]
  \centering
  \resizebox{\columnwidth}{!}{  
\small
\begin{tabular}{l | c c c c}
     \toprule
     \textbf{Model} & \makecell{\textbf{Kendall's $\mathbf{\tau}$ ($\uparrow$)}} & \makecell{\textbf{Spearman's $\mathbf{\rho}$ ($\uparrow$)}} & \makecell{\textbf{Edit dist. ($\downarrow$)}}\\ \midrule
     Chance & 0.0000 & 0.0000 & 6.5822 \\ \midrule
     Repr. Ant. \cite{vondrick2016generating} & 0.3383 & 0.4132 & 5.4596\\ 
     MemDPC \cite{han2020memory} & 0.3492 & 0.4206 & 5.3398  \\
     TAP \cite{tap} & 0.3344 & 0.4107 & 5.4178  \\
 \textbf{\model{} (ours)} & \textbf{0.3632} & \textbf{0.4420} & \textbf{5.3343} \\
         \model{} (vision only)  & 0.3530 & 0.4328 & 5.3370 \\
     \bottomrule
\end{tabular}
}
\vspace{0cm}
\caption{\textbf{Unshuffling image collections:} As defined in Eq.~\ref{eq:putov}, $\log P(u \to v)$ gives a log-likelihood score of $v$ being the future state of $u$. These scores induce a graph which is optimally traversed using a TSP solver. Each model above defines a different $P$ and is applied to shuffled videos from the Recipes test set. Our model outperforms previous state of the art on all metrics.}
\label{tbl:deshuf}
\vspace{-10pt}
\end{table}

\begin{table}[tb]
\centering
\small
\setlength\tabcolsep{4pt}
\resizebox{\columnwidth}{!}{  
\begin{tabular}{cl|ccccc|c}
\toprule
&& \multicolumn{5}{c|}{\textbf{Sampling strategy}} \\
    && \textbf{Rand} & \textbf{Cos sim} & \textbf{TAP} & \textbf{RA} & \textbf{Model} & \textbf{\em Avg} \\ \midrule
    \multirow{6}{*}{\rotatebox[origin=c]{90}{\parbox[c]{1.2cm}{\centering \textbf{Features}}}} & {Random} & 50.4 & 50.7 & 51.3 & 51.4 &51.2 & 51.0\\
     & ImageNet & 51.5& 52.1 & 50.8 & 50.9 &53.4 & 51.7\\
     & Cross-modal & 52.6 & 53.3& 50.9 & 50.6 &55.8 & 52.6\\
     & Repr. Ant. \cite{vondrick2016anticipating} & 50.7 & 51.4 & 51.2 & 51.2 & 51.7 & 51.2 \\
     & TAP \cite{tap} & 50.8 & 51.4 & 51.4 & 51.3 & 51.8 & 51.3 \\
     & \textbf{\model{} (ours)} & 52.3& 53.4 & 50.5 & 50.7 &\textbf{69.2} & \textbf{55.2} \\ \midrule
     & {\em Average} & 51.4 & 52.0 & 51.0 & 51.0 & \textbf{55.5} & 52.2  \\\midrule
     & Chance & 50.0 & 50.0 & 50.0 & 50.0 &50.0 & 50.0\\
     & From scratch  & 51.1 & 52.1 & 51.8 & 51.4 &62.5 & 53.8\\
     \bottomrule
\end{tabular}
}
\vspace{0.2cm}
\caption{\textbf{Learning the arrow of time:} 
We train a linear layer on frozen features as well as a full model from scratch (last row) to detect which of two input frames comes first. This task is near-impossible when sampling random frames. Using our temporal model to sample frames leads to a significant improvement in performance, indicating that our model can identify salient transitions in video. Our visual embedding also outperforms other models, highlighting the importance of temporally-attuned representations.}
\label{tbl:arrowoftime}
\vspace{-1em}
\end{table}

\noindent
\textbf{Discovering the arrow of time:} To further examine whether our model has learned to discover meaningful transitions between states, we explore the arrow of time classification task, introduced in \cite{misra2016shuffle, pickup2014seeing, arrowoftime}. In \cite{arrowoftime}, a network is trained on short videos (on the order of seconds) to predict whether input is being played forward or backward. 

We consider the more challenging task of predicting the temporal relationship between two far-apart input frames -- which one comes first? 
For frames which depict unrelated moments, this task is perhaps near-impossible, even for humans. But if frames show semantically related states, the direction of likely transition provides a useful signal for solving the arrow-of-time task. 

We train linear classifiers on top of frozen features as well as a full network from scratch to solve the arrow of time task on randomly shuffled pairs of frames. We sample pairs of frames using our learned predictive model by selecting the highest-probability futures of start frames selected with the concreteness prior (Eq. \ref{eq:concreteness_cos_prior}). We demonstrate in Table \ref{tbl:arrowoftime} that the temporal ordering of frames mined by our model is much more classifiable than that of frames sampled using predictive models in previous work. Further, our learned features are much more able to classify a given pair of frames, since they must capture temporal information in training. This confirms that a strong understanding of dynamics that emerges from the cycle consistency task. 

\section{Conclusion}
We introduce a self-supervised method to learn temporal dynamics by cycling through narrated video. Despite the simplicity of our architecture, our model is able to discover long-term state transitions in vision and language. We show that this model can be applied without further training to challenging downstream tasks such as anticipating far-away action and ordering collections of image data. 

\textbf{Acknowledgements:} This work was done while Dave Epstein was a student researcher at Google. We thank Alexei Efros, Mia Chiquier, and Shiry Ginosar for their feedback, and Allan Jabri for inspiration in figure design. Dave would like to thank Dídac Surís and Carl Vondrick for insightful early discussions on cycling through time in video.

\renewcommand{\thesubsection}{\Alph{subsection}}
\section{Appendix}

\begin{figure}[H]
\centering
\includegraphics[width=\columnwidth]{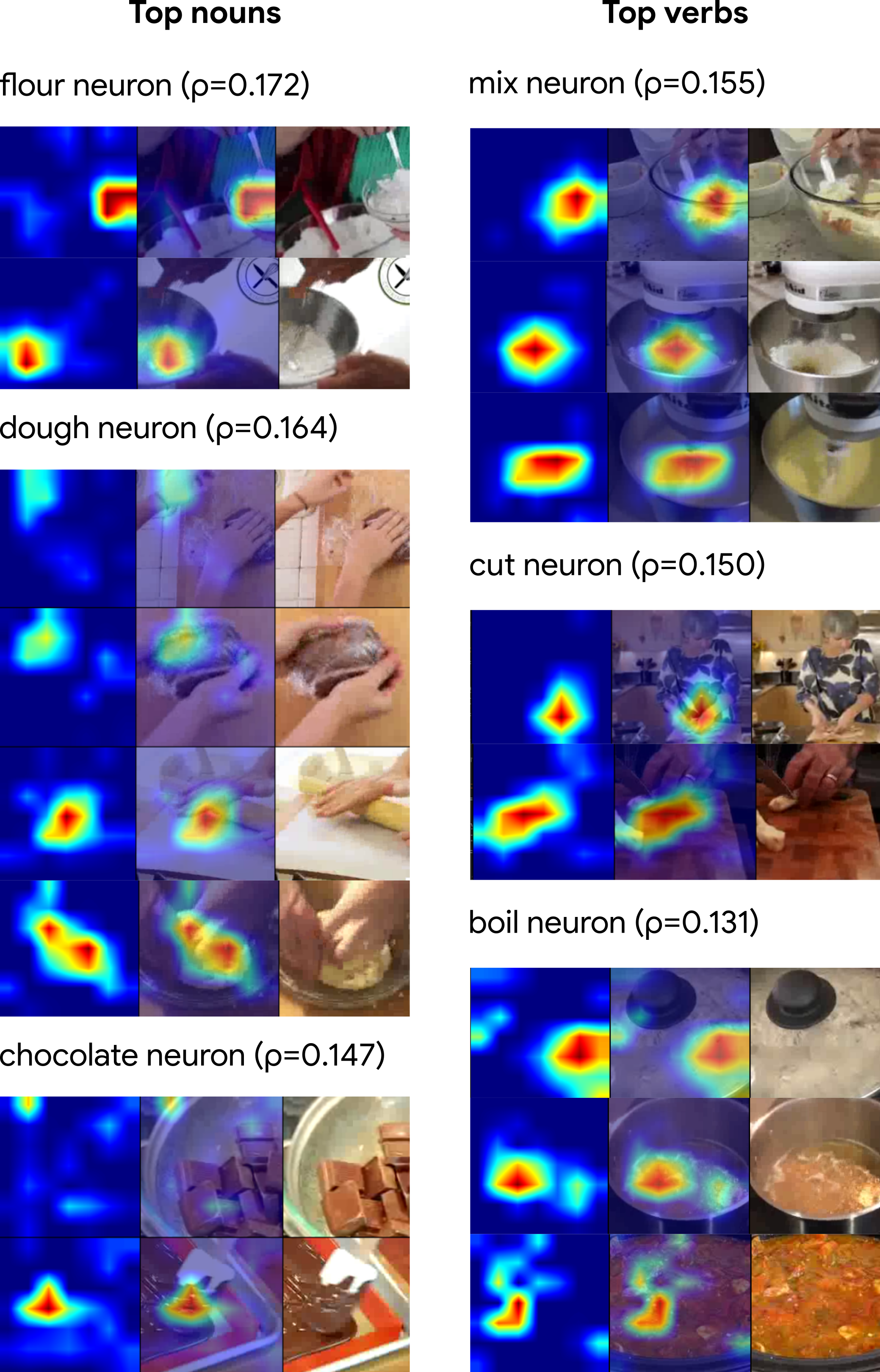}
\caption{\textbf{Dissecting neuron activations:} We find visual neurons that have the highest correlation with words in temporally co-occurring text. We select three verbs and three nouns that are highly correlated to neurons in the representation, and randomly select among the images that most excite these neurons, as representatives of the neuron's function. In each neuron, the left column shows the GradCAM activation of the image, the right column shows the original image, and the center column overlays the two. Despite training only with the indirect supervision of narration, our model discovers many different objects and actions.}
\label{fig:neurons} 
\end{figure}

\subsection{Additional Experiments}
\textbf{Dissecting neuron activations:} To solve our cycle consistency problem, our model must be able to attend to relevant parts of the visual input which inform predictions of the future and past, and lead to strong cross-modal correspondences. We probe our learned representation for discovered objects and actions and visualize this in Figure \ref{fig:neurons}.

First, we compute the correlation between the activation of each neuron in the visual representation given by $\Phi_V$ and the presence of words in corresponding text, using the Spearman rank correlation test run for all $d \cdot |\mathcal{V}|$ neuron-word pairs, where $\mathcal{V}$ is the vocabulary and $d$ the dimension of the embedding. We select neuron-word pairs with the highest correlation score and visualize examples that maximally activate these neurons, along with a GradCAM \cite{gradcam} localization of image regions that caused the activation. We generate the visualization by randomly selecting among images that yield the top 0.01\% of neuron activation values. Note that the images in Figure \ref{fig:neurons} have selected without considering their corresponding textual information, and are filtered only by how much they excite the neuron in question. Our model appears to learn localization of common actions and objects in the training data, despite training without any supervision.

\subsection{Implementation Details}
\subsubsection{Cross-modal correspondence}
A key component of our cycle model is the ability to find correspondences between vision and language. Recent work has demonstrated remarkable progress in training models on massive datasets for this cross-modal retrieval task: given a moment at time $t$ in one modality of a video - $M_t$ - find the matching moment in the other modality $M^\prime$. 

We build on the approach presented in \cite{milnce} which uses a contrastive loss to train representations of vision and language, where temporally co-occurring information is considered ground truth ($M_t$ should retrieve $M^\prime_t$). To handle the common misalignment intrinsic to real-world video, \cite{milnce} allows for representations within $k$ nodes of the ground truth node $M^\prime_t$ to be considered as positives. We modify this approach to improve performance on the case where retrieval must discriminate among a set of candidate moments drawn from the same video as opposed to randomly across the entire dataset. 

The concrete formulation in \cite{milnce}, which shares its general structure with the soft attention formulation used for cycle edges, is:

\begin{equation}
    \mathcal{L}_\text{cross-modal} = -\log\left( \frac{\sum_{i=-k}^k e^{\phi_M(M_t)^T \phi_{M^\prime}(M^\prime_{t+i})}}{\sum_{i,j} e^{\phi_M(M_i)^T \phi_{M^\prime}(M^\prime_{j})} } \right)
\end{equation}

Where $\phi$ are modality-specific embedder networks. We make three modifications to this method which we found to improve performance of cross-modal retrieval:

\begin{enumerate}
    \item We consider a window of $2k + 1$ nodes in both directions of retrieval. We allow both $M_t \to \{M^\prime_{t-k}. \ldots, M^\prime_{t+k}\}$ and $M^\prime_t \to \{M_{t-k}. \ldots, M_{t+k}\}$. This adds a symmetric term to the numerator of the above equation.
    \item We use a Gaussian kernel to weight each positive pair based on the temporal distance between the two moments ($\texttt{weight}(t_i, t_j) \propto \mathbf{P}(\mathcal{N}(0,1) = |t_i - t_j|$), whereas in \cite{milnce} all positive pairs have weight $1$. This adds a weighting term ahead of each summand in the numerator of the above equation.
    \item We require our cross-modal retrieval to successfully compute correspondences among many candidates from the same video. Therefore, we augment the set of negatives with many moments from the same target video, as opposed to \cite{milnce} which randomly samples negatives from throughout the dataset. This encourages the model to distinguish between moments using more fine-grained temporal cues and fewer higher-level topic cues.
\end{enumerate}
\subsubsection{Training}
Because good cross-modal correspondence is necessary to learn strong, semantic cycles, we initialize $\lambda_2 = 1$ and exponentially increase $\lambda_1$ from some small value $\epsilon$ up to a final value in $[0.5, 2]$ (depending on the experiment), across 30 epochs. We peg $\lambda_3$ to be a fixed ratio of $\lambda_1$ when using the similarity loss -- specifically, we set $\lambda_3 = 3 \lambda_1$. We find that this schedule maintains high performance on the cross-modal correspondence task even when updating model weights to solve cycles. We train with learning rate $0.0001$ with the Adam optimizer \cite{Adam} until convergence.

We set ${p_\text{unimodal} = 0.5}$ to balance between leveraging cross-modal information and learning to function on only one modality. Cycle edges and the cycle loss are computed within each video separately, whereas the cross-modal loss uses negatives from across the entire batch. 

For the text architecture, each word in an input utterance is first mapped to a vector in $\mathbb{R}^d$ by running through an embedding matrix $W \in \mathbb{R}^{\texttt{num words} \times d}$ and another linear layer $w_1 : \mathbb{R}^d \to \mathbb{R}^d$. These vectors are then run through a ReLU and max-pooled to yield a single vector in $\mathbb{R}^d$, which is run through a layer $w_2 : \mathbb{R}^d \to \mathbb{R}^d$ to give the final utterance embedding. 
\subsection{Effect of loss terms}
Our model relies on cross-modal correspondence to use information from both modalities, but this can be learned in separate pre-training stages, which allows $\lambda_2 = 0$. This decreases cycle accuracy by 7\%. Co-training with this loss term (and annealing its weight, Sup. Mat. L152-156) prevents catastrophic forgetting of cross-modal correspondences, which are used to learn cycles. The visual similarity loss is optional, since the constrained temporal attention functions well on its own, so we can set $\lambda_3 = 0$, slightly impacting performance ({\em e.g.}, -1\% on cycle rank). The model does not learn temporal dynamics at all with $\lambda_1 = 0$; this can be thought of as the cross-modal baseline. 

{\small
\bibliographystyle{ieee_fullname}
\bibliography{egbib}
}

\end{document}